%% file: main.tex
\documentclass[a4paper,fleqn]{cas-dc}

\usepackage[numbers]{natbib}
\usepackage[small]{caption}
\usepackage{graphicx}
\usepackage{amsmath}
\usepackage{amsthm}
\usepackage{booktabs}
\usepackage{algorithm}
\usepackage{algorithmic}
\usepackage{epsfig}
\usepackage{amssymb}

\usepackage{multirow}
\usepackage{subcaption}

\usepackage{tikz}
\usepackage{comment}

\newcommand{\Skip}[1]
{
}

\makeatletter
    \let\@internalcite\cite
    \def\cite{\def\citeauthoryear##1##2{##1, ##2}\@internalcite}
    \def\shortcite{\def\citeauthoryear##1{##2}\@internalcite}
    \def\@biblabel#1{\def\citeauthoryear##1##2{##1, ##2}[#1]\hfill}
\makeatother

\def\tsc#1{\csdef{#1}{\textsc{\lowercase{#1}}\xspace}}
\tsc{WGM}
\tsc{QE}

\begin{document}
\let\WriteBookmarks\relax
\def\floatpagepagefraction{1}
\def\textpagefraction{.001}

\shorttitle{}    

\shortauthors{}  

\title [mode = title]{Boundary-Recovering Network for Temporal Action Detection}  

\author[1]{Jihwan Kim}
\author[3]{Jaehyun Choi}
\author[1]{Yerim Jeon}
\author[1,2]{Jae-Pil Heo\corref{cor1}}
\affiliation[1]{organization={Department of Artificial Intelligence, Sungkyunkwan University},
            city={Suwon},
            postcode={16419}, 
            country={South Korea}}

\affiliation[2]{organization={Department of Computer Science and Engineering, Sungkyunkwan University},
            city={Suwon},
            postcode={16419}, 
            country={South Korea}}
\affiliation[3]{organization={Department of Electrical Engineering, Korea Advanced Institute of Science and Technology},
            city={Daejeon},
            postcode={34141}, 
            country={South Korea}}

\cortext[cor1]{Corresponding author at: Department of Artificial Intelligence, Sungkyunkwan University, Suwon, 16419, South Korea. E-mail addresses: damien911224@gmail.com~(J. Kim), PRE6ENT@gmail.com~(J. Choi), jyr990330@gmail.com~(Y. Jeon), jaepilheo@skku.edu~(J.-P. Heo)}

\input{sec/0_abstract}



\maketitle

\input{sec/1_introduction}
\input{sec/2_related_work}
\input{sec/3_our_approach}
\input{sec/4_experiments}
\input{sec/5_conclusion}








\appendix

\printcredits

\bibliographystyle{cas-model2-names}

\bibliography{bibliography}



\end{document}

%% file: sec/0_abstract.tex
\begin{abstract}
Temporal action detection (TAD) is challenging, yet fundamental for real-world video applications.
Large temporal scale variation of actions is one of the most primary difficulties in TAD.
Naturally, multi-scale features have potential in localizing actions of diverse lengths as widely used in object detection.
Nevertheless, unlike objects in images, actions have more ambiguity in their boundaries.
That is, small neighboring objects are not considered as a large one while short adjoining actions can be misunderstood as a long one.
In the coarse-to-fine feature pyramid via pooling, these vague action boundaries can fade out, which we call `vanishing boundary problem'.
To this end, we propose Boundary-Recovering Network (BRN) to address the vanishing boundary problem.
BRN constructs scale-time features by introducing a new axis called scale dimension by interpolating multi-scale features to the same temporal length.
On top of scale-time features, scale-time blocks learn to exchange features across scale levels, which can effectively settle down the issue.
Our extensive experiments demonstrate that our model outperforms the state-of-the-art on the two challenging benchmarks, ActivityNet-v1.3 and THUMOS14, with remarkably reduced degree of the vanishing boundary problem.
\end{abstract}

%% file: sec/1_introduction.tex
\section{Introduction}

\begin{figure}[t]
\centering
\includegraphics[width=8.35cm]{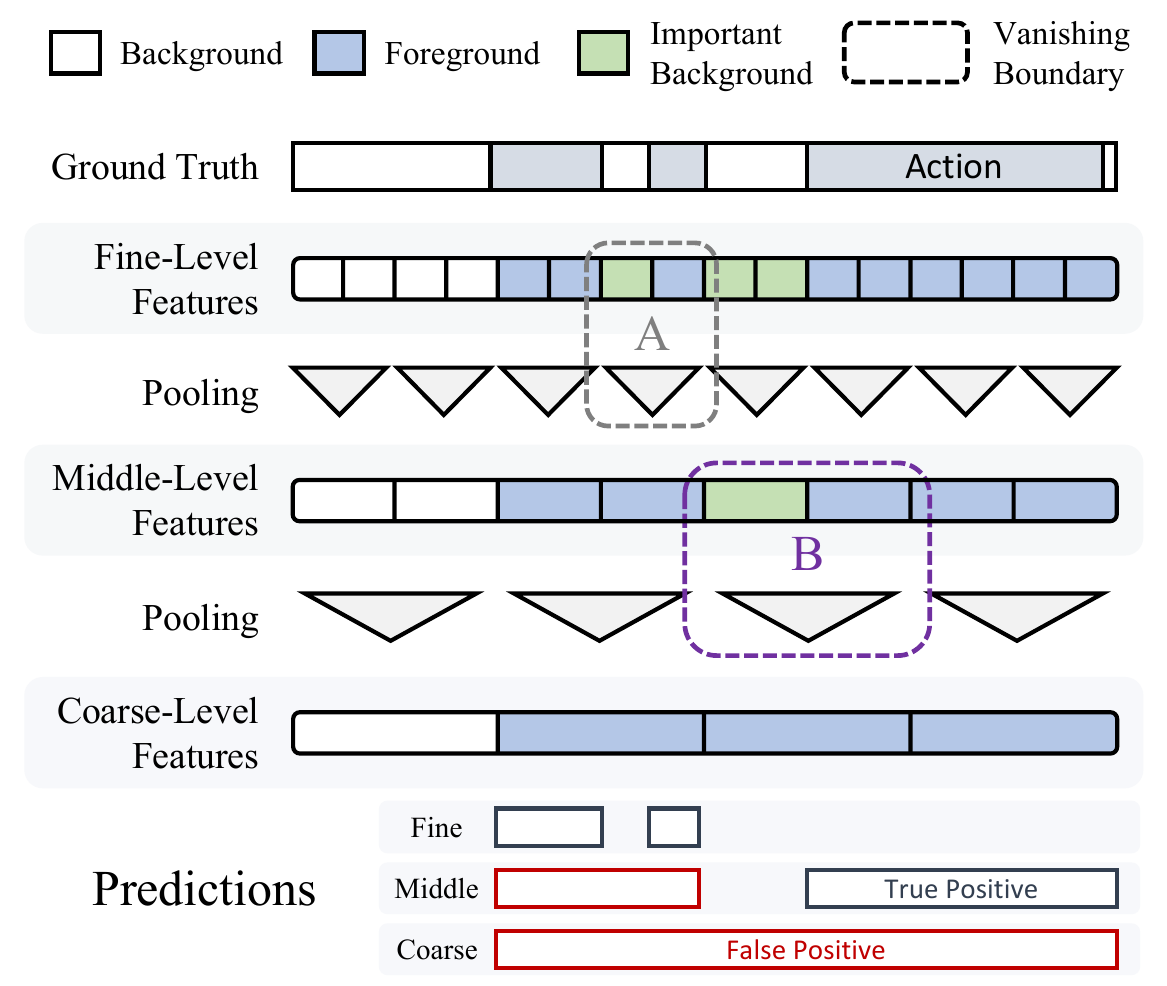}
\caption{\textbf{Illustration of the vanishing boundary problem.}
In our intuitive example, three action instances go through coarse-to-fine pyramid network with pooling operations.
The features of background frames can easily dim out via pooling due to absence of clear patterns related to the action.
However, this naive pooling process introduces vanishing boundary problem when the backgrounds exist between short action instances, labeled as important background like the points of `\textbf{A}' and `\textbf{B}'.
As a result, the problem can cause the model to predict long false positives in coarser-levels due to temporal ambiguity of action boundaries.
} 
\label{fig:introduction}
\end{figure}

The need for understanding videos has gained great attention as an innumerable amount of videos got published on online platforms.
Initially, the field started with video classification using trimmed video clips as input in order to train models to correctly classify actions.
However, cost of snipping the real-world videos for classification triggered the literature towards Temporal Action Detection (TAD).
Temporal action detection aims to not only correctly classify actions but also identify temporal boundaries in untrimmed videos.

Pioneering methods~\cite{escorcia2016dap, caba2016fast, buch2017sst} deployed fixed-length windows called action proposals inspired by the field of object detection~\cite{girshick2014rcnn, girshick2015fast-rcnn, ren2015faster-rcnn}.
The following approaches~\cite{zhao2017ssn, yuan2017maximul-sum, lin2018bsn} have introduced point-wise learning where they predict probabilities of start and end boundaries at each time step to cope with the low-recall issue.
Based on these fine-grained predictions, they generate flexible action proposals by grouping each pair of start and end boundaries.
Although they have achieved high-recall performance, a bunch of generated proposals with various lengths make the ranking process more challenging.
Recently, many models have been devised for precise temporal action detection focusing on a more accurate ranking.

The challenge of large temporal scale variation stands out as one of the factors behind the difference in performance.
Although actions have a large variation of scales, the previous approaches utilize single-scale features while multi-granularity features are widely used in object detection~\cite{lin2017retina, tian2019fcos} to effectively handle various object scales.
Feature pyramid network (FPN)~\cite{lin2017fpn}, one of the most popular multi-scale architectures, constructs several levels of spatial granularity by combining high-level semantics and low-level details.
In this multi-scale framework, each level of features is responsible for object instances in the corresponding spatial span.
This effectively eases the difficulty of the task and reduces the imbalanced performances over scales, especially in preventing over-fitting to large objects.
With these benefits of multi-scale features, several research attempts~\cite{long2019gtan, liu2020progressive, lin2021salient, zhao2021stitching} have been recently made in temporal action detection.

However, there is a clear difference between objects and actions for multi-scale features.
Unlike shapes of objects, actions are distinguished heavily depending on their surrounding temporal context.
That is, an action instance is identified by background frames on both side of the boundaries, which define the start and end of the action.
On top of that, actions with a short interval of background or repeating motions can be naturally grouped as a long one in annotations while neighboring or even overlapped objects are never considered as a large one in labels.
Therefore, if we remove the background clip between two actions, it is highly possible to view the actions as a long instance even for humans.
When this temporal ambiguity of action boundaries meets feature pyramid architecture, the problem could cancel out the benefits of the multi-scale features since crucial boundary representations can be pooled out.

We define this problem as `vanishing boundary problem' as shown in Fig.~\ref{fig:introduction}.
In the coarse-to-fine feature pyramid, the features go through several pooling operations.
In this procedure, important boundary features can easily fade out since the background frames do not show any consistent patterns with regard to the actions.
This results in unexpected false positives on coarser levels since the model falsely views the actions as a single longer one.
Hence, when we directly deploy the feature pyramid architecture for temporal action detection, the above issue could significantly reduce the merits of multi-scale features, which is essential for handing various scales of actions.

To settle down the aforementioned issues, we propose a new framework, Boundary-Recovering Network (BRN), which introduces scale-time blocks with scale-time features.
Firstly, scale-time features are the template to allow for scale-time blocks to learn over scales to recover boundary patterns dimmed out by pooling.
In order to build scale-time features, multi-scale features are simply interpolated to the same temporal length.
Then we stack them on a new scale dimension axis to construct scale-time features.
These features enable BRN to learn representations over scales just like it does on time axis.
Secondly, on top of the scale-time features, scale convolutions and time convolutions in the scale-time blocks aggregate features across the multi-granular levels.
More specifically, scale convolutions directly exchange features with other levels when important boundary information is missed at a coarse level.
The multi-rate dilated convolutions with selection module further allows for fine-grained feature exchange among neighboring scales or distant levels.
In this way, scale-time blocks are designed to relieve the vanishing boundary problem.

Our contributions are summarized as follows:
\begin{itemize}
	\item To the best of our knowledge, it is the first approach to shed light on the problem of multi-scale features caused by the temporal ambiguity in action boundaries, called `vanishing boundary problem'.
	\item We propose a novel framework, Boundary-Recovering Network (BRN), which introduces scale-time blocks with scale-time features to handle the problem of missing important boundary information during the pooling process.
	Scale-time blocks learn to fuse features across the scale levels, which effectively reduces the effect of the vanishing boundary problem.
	\item Our extensive experiments demonstrate that the proposed framework outperforms the existing state-of-the-art methods on the two challenging benchmarks, ActivityNet-v1.3 and THUMOS14.
\end{itemize}

%% file: sec/2_related_work.tex
\section{Related Work}

\subsection{Temporal Action Detection}
Early methods~\cite{zhao2017ssn, chao2018tal-net, gao2018ctap, kim2019coarsefine} have achieved great advance in TAD during the last decade. 
Recently, point-wise learning has been widely introduced to generate more flexible action proposals without pre-defined temporal windows.
BSN~\cite{lin2018bsn} and BMN~\cite{lin2019bmn} grouped candidate start-end pairs to generate action proposals, then ranked them for final detection outputs.
BSN++~\cite{su2021bsn++} tackled scale imbalance problem based on BSN.
Besides, graph neural networks are getting a great deal of attention in the field of temporal action detection~\cite{xu2020gtad, zeng2019pgcn}.
PGCN~\cite{zeng2019pgcn} improved the ranking performance via constructing a graph of proposals based on their overlaps. 
GTAD~\cite{xu2020gtad} considered temporal action detection as a sub-graph localization problem and proposed a new framework with graph neural network.

On the other hand, CSA~\cite{sridhar2021class} proposed class-semantic based attention, which can be applied to many TAD methods and showed a clear improvement upon the baseline.
ContextLoc~\cite{zhu2021contextloc} introduced a new framework to enrich local and global contexts for TAD.
Also, RTD-Net~\cite{tan2021relaxed} devised an architecture based on Transformer~\cite{vaswani2017attention} for direct action proposal generation.
ActionFormer~\cite{zhang2022actionformer} and Self-DETR~\cite{kim2023self} deployed transformer encoder as backbone network.
TAGS~\cite{nag2022tags} introduced proposal-free mask learning for TAD.

\subsection{Multi-Scale Features}
As multi-granularity features are widely used in object detection~\cite{lin2017fpn, lin2017retina, tian2019fcos}, there has been methods with multi-granularity architecture in temporal action detection.
GTAN~\cite{long2019gtan} learns Gaussian kernels to predict the length of action instances on multi-scale features with an one-stage architecture.
MGG~\cite{liu2019mgg} is based on the U-Net~\cite{ronne2015unet} architecture to create multi-scale features.
They also combined the anchor-based and anchor-free (point-wise) methods.
PBRNet~\cite{liu2020progressive} focused on the multi-stage architecture for iterative refinement with U-NET-like architecture.
Moreover, AFSD~\cite{lin2021salient} adopted an anchor-free detection framework with FPN and introduced a refinement module with boundary pooling.
VSGN~\cite{zhao2021stitching} pointed out the imbalanced performances of scales and devised a way of augmentation to focus on short action instances and deployed a multi-scale architecture with graph neural network.

Although our framework is based on a multi-granularity architecture, our work is different from the prior arts by specifically focusing on the vanishing boundary problem and proposing novel viewpoint, scale-time representation.

%% file: sec/3_our_approach.tex
\begin{figure*}[t]
\centering
\includegraphics[width=17.35cm]{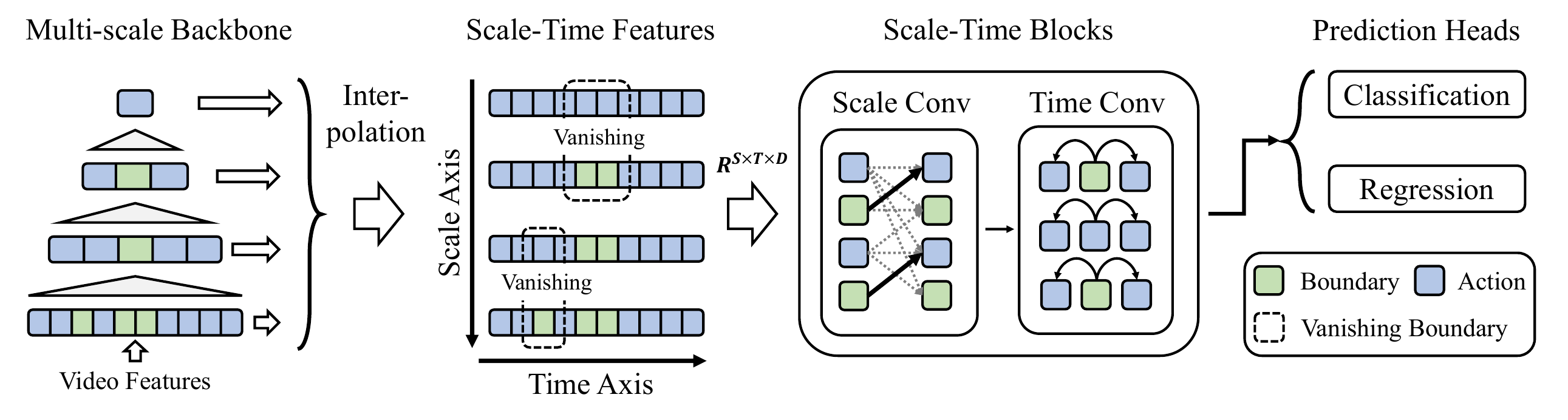}
\caption{\textbf{Overall architecture of Boundary-Recovering Network (BRN).}
First, the features for a video from a pre-trained 3D CNN are fed into the backbone network to construct multi-scale features.
Second, simple interpolation builds scale-time features. 
Finally, the scale-time blocks learn to exchange features over scales to recover the boundary patterns.}
\label{fig:architecture}
\end{figure*}

\section{Our Approach}
To settle down the vanishing boundary problem, we propose a new framework, Boundary-Recovering Network (BRN), which introduces scale-time features and scale-time blocks as illustrated in Fig.~\ref{fig:architecture}. 

BRN introduces scale-time representations on a one-stage detection scheme with multi-scale features.
It has the following main components: 1) backbone network constructing multi-scale features, 2) scale-time representations (features and blocks), and 3) prediction heads.
We first elaborate the architecture of the backbone network.
Then we introduce the details of scale-time features and scale-time blocks.
Finally, the objectives are defined in the following section with the details of the prediction heads.

\subsection{Backbone Network}
\noindent\textbf{Base Features.} As the input to the overall framework, we use the features of a 3D CNN pre-trained on Kinetics~\cite{kay2017kinetics}, and fixed while training our model.
To extract features, each video is fed into the 3D CNN, and further global-average pooled in spatial dimensions so that only the temporal dimension remains.
Formally, the input of our model for a video $v$ is described as follows:
\begin{eqnarray}
F_v=\{f_1,f_2,...,f_{L_v}\},
\end{eqnarray}
where $f_i$ is a temporal feature, and $L_v$ is the length of the features for a video $v$, which is $1/8$ frame length of the video $v$ in our framework.

\vspace{2pt}
\noindent
\textbf{Multi-Granularity.} 
After extracting temporal features with the pre-trained 3D CNN, multi-scale representations are constructed with the backbone network.
In the backbone network, each layer has a pooling layer, which shrinks the temporal resolution of the features with a stride of 2.
Let us denote the outputs of each layer in the backbone network as follows:
\begin{eqnarray}
B = \{B_1, B_2, ..., B_S\},
\end{eqnarray}
where $B_i$ is the outputs of the $i$-th layer of the backbone network, and $S$ is the number of multi-scale levels.

In our framework, we can build multi-scale features from two types of the backbone network: 1) convolution, 2) transformer layers.
Both types are used in many methods in TAD.
Our methods can be applied to various architectures deploying multi-scale features.
In this paper, we will demonstrate this merit in the experiments with both types of the backbone network.

\Skip{
These multi-scale features are then fed into Bidirectional Feature Pyramid Network (BiFPN) inspired by EfficeintDet~\cite{tan2020efficientdet}.
BiFPN has top-down and bottom-up streams to learn high-level semantics with low-level details.
Specifically, we can denote the input, intermediate, and output features of BiFPN as $P^{\text{In}}_i$, $P^{\text{Mid}}_i$, and $P^{\text{Out}}_i$, respectively, where $P_i$ stands for the features of the $i$-th level.
Then the intermediate and output features can be represented as below:
\begin{eqnarray}
{\small
\begin{aligned}
P^{\text{Mid}}_i &= \text{Conv}(\frac{w_1 \cdot P^{\text{In}}_i + w_2 \cdot \uparrow_{2.0}(P^{\text{In}}_{i + 1})}{w_1 + w_2 + \epsilon}), \\
P^{\text{Out}}_i &= \text{Conv}(\frac{w'_1 \cdot P^{\text{In}}_i + w'_2 \cdot P^{\text{Mid}}_i + w'_3 \cdot \uparrow_{2.0}(P^{\text{Out}}_{i - 1})}{w'_1 + w'_2 + w'_3 + \epsilon}),
\end{aligned}
}%
\label{eq:bifpn}
\end{eqnarray}
where $\text{Conv}(\cdot)$ means 1D temporal convolution with a kernel size of 3, and $\uparrow_{s}(\cdot)$ is linear interpolation for upsampling the length $s$ times. 
Also, $w_i$ and $w'_i$ are learnable scalar parameters as the weights of fusion, and $\epsilon$ is a small scalar to prevent zero division.
Note that Eq.~\ref{eq:bifpn} is for $P_2$, $P_3$, and $P_4$.
That is, there are no $P^{\text{Mid}}_1$ and $P^{\text{Mid}}_5$ for computational efficiency as done in ~\cite{tan2020efficientdet}.
}

\subsection{Scale-Time Representations}
\noindent\textbf{Motivation.} 
The vanishing boundary problem in TAD leads to the loss of boundary information, particularly on coarse levels, when using multi-scale features. 
When we forward these multi-scale features to prediction heads, it will cause many long false positives as they can easily consider neighboring short actions as a long one.
To address this issue, we propose scale-time representations.
The goal is to recover the missing boundary patterns by using scale-time blocks, which facilitate the exchange of features over different scales.
We first interpolate multi-scale features into scale-time features, aligning the time axis over scales to enable precise and fine-grained feature aggregation.
Then the scale-time blocks, based on multi-rate dilated convolutions with an attention module, learn to bring features from the appropriate scale level at each time step, effectively recovering the missing boundary patterns.
From this design, we effectively relieve the vanishing boundary problem.

\vspace{2pt}
\noindent
\textbf{Scale-Time Features.} 
To relieve the vanishing boundary problem, we introduce Scale-Time Features (STF) and Scale-Time Blocks (STB).
We firstly present scale-time features, which is a new viewpoint of coarse-to-fine features for temporal action detection. 
Multi-scale features can form a new \textbf{scale} axis by interpolating features of each scale level into the same temporal length. 
We define these interpolated features as scale-time features.
Although features of each scale have the same length in STF, their granularity is different in themselves, which means that they are still suitable for detecting action instances for the corresponding scale.

Specifically, the output $B_i$ of each scale level first goes through a 1D temporal convolution layer with a kernel size of 1 to embed $B_i$ so that they have the same number of channels.
Then, they are upsampled to temporal length of $T$ with linear interpolation.
Specifically, the interpolation of features is conducted independently on each channel along the time dimension, and works in the same way as the bi-linear upsampling of images.
Let us define $\text{STF}_i$ as the scale-time features of the $i$-th level of scale.
Finally, scale-time features can be represented as below:
\begin{eqnarray}
\begin{aligned}
\text{STF}_i &= \text{Resize}(\text{Conv}(B_i), T), \\
\text{STF} &= \text{Stack}(\text{STF}_1, \text{STF}_2, ..., \text{STF}_S),
\end{aligned}
\end{eqnarray}
where $\text{Resize}(\cdot, T)$ is a upsampling function with linear interpolation with the target size of time $T$, and $\text{Stack}(\cdot)$ is stacking the interpolated features on a new scale axis. Also, $S$ is the number of multi-scale levels.
Now, STF is a tensor of $R^{S \times T \times D}$, where $D$ is the number of the channels.

\begin{figure}[t]
\centering
\includegraphics[width=8.35cm]{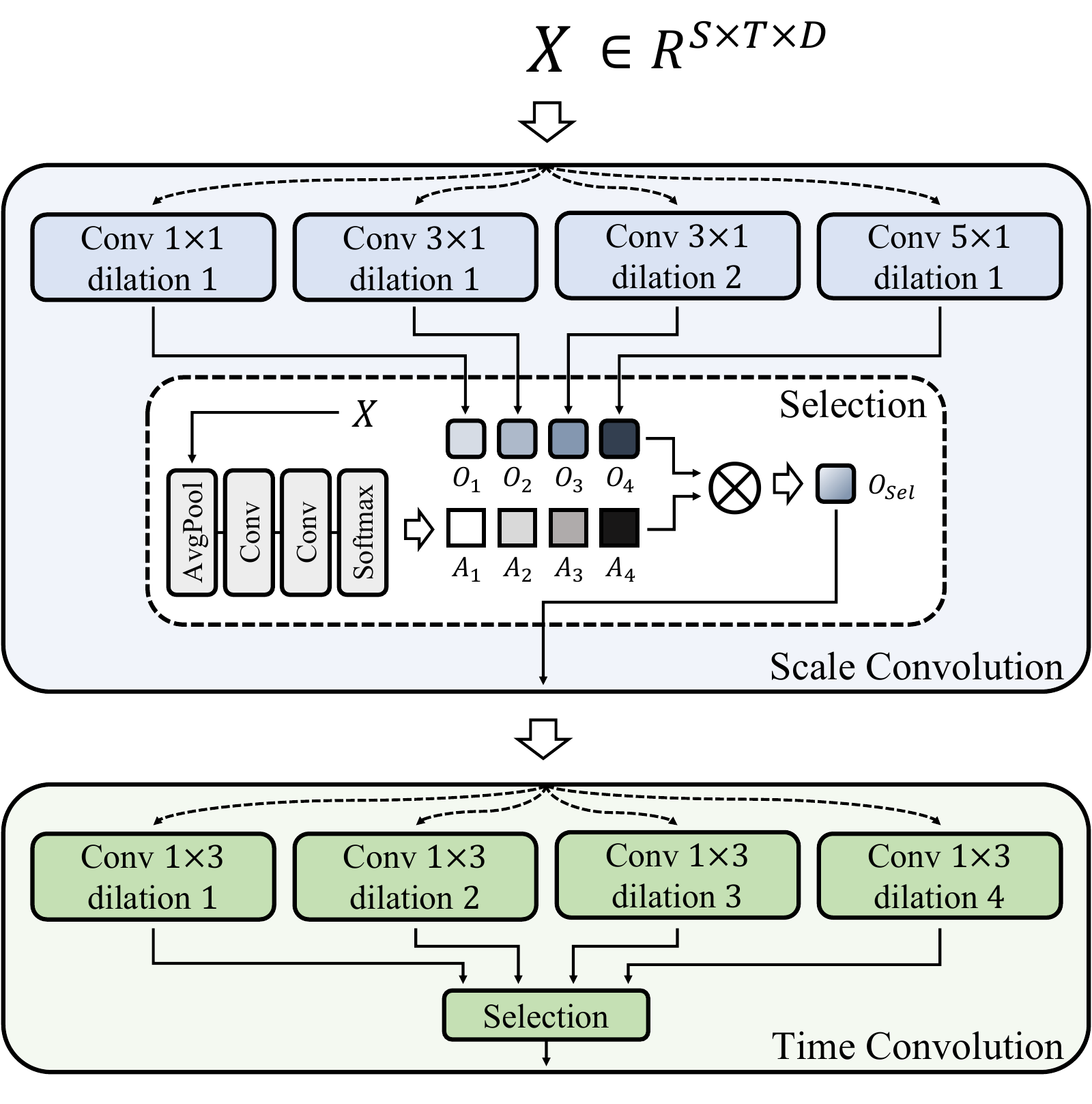}
\caption{\textbf{Scale-Time Blocks.} 
Scale and time sub-blocks have dilated convolutions with different rates and kernel sizes. 
Afterwards, outputs $O_i$ are aggregated with a selection module, an attention-based pooling as in Eq.~\ref{eq:STB_GAP}, \ref{eq:STB_ATT}, \ref{eq:STB_Out}.
}
\label{fig:ST_block}
\end{figure}

\vspace{2pt}
\noindent
\textbf{Scale-Time Blocks.} After forming scale-time features, Scale-Time Blocks (STB) learn scale and time representations on these features.
STB has two main components; scale and time convolution sub-blocks as shown in Fig.~\ref{fig:ST_block}. 
The scale convolution sub-block has dilated convolutions with multiple different dilation rates and kernel sizes.
This kind of architecture is widely used in semantic segmentation to represent features with various receptive field sizes.

Different kernel sizes and dilation rates of convolutions can act as selection of bringing features from neighboring levels or distant levels.
In this way, the model can learn to handle the degree of feature exchange across scales via selection weights of the multiple outputs.
For example, the $1 \times 1$ convolution does not bring features from other levels while the $3 \times 1$ or $5 \times 1$ convolutions aggregate representations from other scales with different ranges.
At certain time steps, the selection weights of the outputs with the $1 \times 1$ kernel will be high if there is no need for the features from other scale levels.
Similarly, when the information of other levels is needed like in the case of vanishing boundary problem, the weights of $3 \times 1$ or $5 \times 1$ should be high.

Formally, let us denote the input of the sub-block as $X$ and the four kinds of outputs as $O_1, O_2, O_3$ and $O_4$, respectively, which are in $R^{S \times T \times D}$.
Now we aim to weigh each of $O_i$ differently as they are from convolutions with various kernel sizes or dilation rates for different degree of feature exchange.
First of all, we form aggregated output by average pooling the input of the block $X$ along with the scale axis:
\begin{eqnarray}
X_{\text{Agg}} = \text{AvgPool}(X),
\label{eq:STB_GAP}
\end{eqnarray}
where $\text{AvgPool}(\cdot)$ is average pooling on the scale axis with a stride of 1 and the kernel size of $5$.
Note that we pad the input $X$ so that $X_{\text{Agg}}$ has the same shape of $X \in R^{S \times T \times D}$.

Then we use $X_{\text{Agg}}$ to produce selection weights for $O_i$ via two $1 \times 1$ convolution layers:
\begin{eqnarray}
A = \text{Conv}_{D \rightarrow 4}~(\text{ReLU}~(\text{Conv}_{D \rightarrow D}~(X_{\text{Agg}}))),
\label{eq:STB_ATT}
\end{eqnarray}
where $\text{Conv}_{D_{\text{in}} \rightarrow D_{\text{out}}}$ means a $1 \times 1$ convolution layer with the input channel $D_{\text{in}}$ and the output channel $D_{\text{out}}$.

$A$ is then further reshaped and transposed to $R^{S \times T \times 1 \times 4}$. 
Finally, softmax function is applied to produce selection weights for aggregation.
Then the final outputs of the selection module are described as follows:
\begin{eqnarray}
O_{Sel} = \sum_{i=1}^{4}A_i \otimes O_i,
\label{eq:STB_Out}
\end{eqnarray}
where $A_i \in R^{S \times T \times 1}$ is the selection weights for the $i$-th convolution outputs $O_i$, and $\otimes$ is an element-wise multiplication.

In this way, the model can learn to select appropriate feature levels to exchange which boosts the capability to tackle the vanishing boundary problem. 
As for time sub-block, the selection module is similarly defined to that of scale convolutions.
The time convolution sub-block also has multi-rate dilation convolutions with different sizes of kernels.
The selection module pools and aggregates the features along the time axis instead of the scale axis.

\subsection{Objectives}
\noindent\textbf{Prediction Heads.}
The features obtained through scale-time representations are fed into two different heads: classification and regression heads.
The classification head predicts probabilities of the action categories for each time step, whereas the regression head outputs the distances between the boundaries and the current time step.

Specifically, classification and regression heads both have three convolution layers followed by linear prediction. 
Each convolution layer has scale and time convolutions with a kernel size of 3.
Finally, classification and regression heads go through softmax and sigmoid functions, respectively, for the final predictions.
Note that, our classification head also predicts the probability of background class.

\vspace{2pt}
\noindent
\textbf{Cost Functions.} Let us denote $M_v$ ground-truth action instances for a video $v$ as $Y^*_1,...,Y^*_{M_v}$ and each instance $Y^*_k$ as a pair $(I^*_k, C^*_k)$, where $I_k$ is a normalized time interval $I^*_k=[ t^s_k , t^e_k ] \in [0, 1]$, and $C^*_k \in \{1,2,...,K\} $ is an action class when we have $K$ action categories in consideration. 
Note that $t^s_k$ and $t^e_k$ are the start and end times of the $k$-th action instance, respectively.

Then for the corresponding ground-truth action instances, we can define the full objectives of the framework as follows:
\begin{eqnarray}
\mathcal{L} = l_{cls} + \lambda l_{reg},
\end{eqnarray}
where $l_{cls}$, $l_{reg}$ are the classification and regression cost functions, and $\lambda$ is the weight of the regression loss.

Also, let us denote the predictions of classification and regression as $p_{c}$ and $p_{r}$, respectively.
The classification loss is based on the focal loss~\cite{lin2017retina} defined as below:
\begin{eqnarray}
l_{cls} = -(1 - p_c^{C^*_k})^{\alpha}\log{p_c^{C^*_k}},
\label{eq:loss_cls}
\end{eqnarray}
where $p_c^{C^*_k}$ is the probability of the ground-truth class $C^*_k$ for the instance $Y^*_k$ or that of the background class, and $\alpha$ is the weight of the focal loss. Moreover, the regression loss for a positive proposal is defined as following:
\begin{eqnarray}
l_{reg} = 1 - \text{IoU}(I^*_k, p_r),
\label{eq:loss_reg}
\end{eqnarray}
where IoU is Intersection-over-Union score between ground-truth time interval $I^*_k$ and regression prediction $p_r$. 
Note that positive instances are defined when the time step is within an action instance. 
If multiple ground-truth instances are matched, the one with the highest IoU is chosen for loss.

%% file: sec/4_experiments.tex
\section{Experiments}
\subsection{Implementation Details}
Our experiments are conducted on the two challenging benchmarks of temporal action detection: THUMOS14~\cite{jiang2014thumos14} and ActivityNet-v1.3~\cite{caba2015activitynet}.
We use the features of I3D pre-trained on Kinetics.
We follow video pre-processing process as in~\cite{carreira2017i3d}, which uses 25 FPS for frame extraction and resizes the frames so that they have 256 pixels in their longer side (width or height).
In order to extract optical flow, we use the TVL1 algorithm.
After pre-training the I3D, we fixed the weights of I3D and extract temporal features with 64-frame windows, and use the final convolution outputs with spatial global average pooling.
Each temporal feature is corresponding to 8 frames since the final temporal resolution of I3D is $1/8$ of the input video. 

\vspace{3pt}
\noindent\textbf{Architectures. }
We implement BRN based on two types of models, FCOS and ActionFormer~\cite{zhang2022actionformer}.
FCOS indicates temporal version of the FCOS~\cite{tian2019fcos} model in object detection, which is similar to the backbone of AFSD~\cite{lin2021salient}.
Our scale-time representations can be generic for the multi-scale backbone.
In order to validate the benefits of our framework, we evaluate BRN based on these two types of the backbone networks.

The FCOS backbone is a temporal version of the FCOS~\cite{tian2019fcos} model in object detection, which is similar to the backbone of ASFD~\cite{lin2021salient}.
Specifically, the backbone network has $L$ layers, and each layer has one convolution block and a max-pooling layer with a stride of 2.
Therefore, we have the multi-scale features of $L$ scale levels.
The FCOS backbone also has a projection layer to reduce the dimensions of the input video features to $256$, which is the hidden dimensions of the FCOS backbone.

The ActionFormer backbone is the same as the ActionFormer~\cite{zhang2022actionformer} model.
It also has $L$ layers, and each layer has a transformer block followed by a max-pooling with a stride of 2, producing $L$-level multi-scale features.
We follow the model configuration of the original version as in ~\cite{zhang2022actionformer}.

\vspace{2pt}\noindent
\textbf{Training.}
As for both datasets, we use Adam as the optimizer with the batch size of 16.
For the input, we use 256 and 192 lengths of temporal features for the FCOS and ActionFormer backbones, respectively.
Moreover, we randomly crop the temporal features for data augmentation.
As for the ActionFormer backbone, we follow the training configuration as in \cite{zhang2022actionformer}.

The FCOS backbone is trained by the following configuration.
In THUMOS14, we train the framework for 1200 epochs. The learning rate is decayed by $1/10$ when it reaches 1000 and 1100 epochs.
Also, we randomly slice the video features with a 256-length window.
As for ActivityNet-v1.3, 120 epochs are taken for training. The learning rate decreases by $1/10$ at 80 and 100 epochs.
In addition, we resize the features of a video with linear interpolation to 256.

Moreover, we do not use the concept of `centerness' in FCOS~\cite{tian2019fcos} since the center of an action is ambiguous.
Therefore, we do not employ center sampling which only defines points near the center of an object as positive samples.
Instead, we use all time steps within the time intervals of the ground-truth actions as positive samples.
As for the focal loss in the objective, we use $\alpha = 4$ for the focal loss in $L_{cls}$ defined in Eq.~9 of the paper.

\vspace{2pt}\noindent
\textbf{Inference.} 
We slice the temporal features with a 256-length window with overlap of 128 for the FCOS backbone in THUMOS14.
The ActionFormer backbone receives full-length video features as input as in \cite{zhang2022actionformer}.
As for ActivityNet-v1.3, we resize the features to 256 and 192 lengths as done in training for the FCOS and ActionFormer backbones, respectively. 
Also, we use the top 100, and 200 predictions after non-maximum suppression (NMS) for the final localization results for ActivityNet-v1.3, and THUMOS14, respectively for the both backbones.
We use the IoU threshold of $0.65$ for ActivityNet-v1.3, and $0.50$ for THUMOS14 for NMS in the FCOS backbone.
As for the ActionFormer backbone, we follow the inference set-up as in \cite{zhang2022actionformer}.
For the class label, we fuse our classification scores with the top-1 video-level predictions of \cite{zhao2017cuhk} as done in ~\cite{lin2018bsn,lin2019bmn,xu2020gtad} for ActivityNet-v1.3 in the both backbones. 

\subsection{Comparison with the State-of-the-Art}
We compare the-state-of-the-art methods to evaluate our framework on THUMOS14 and ActivityNet-v1.3 datasets with various baselines.
Our framework has a similar aspect to PBRNet~\cite{liu2020progressive}, AFSD~\cite{lin2021salient}, VSGN~\cite{zhao2021stitching}, and ActionFormer~\cite{zhang2022actionformer} in respect to deploying the multi-scale architecture.

\begingroup
\setlength{\tabcolsep}{9.40pt} 
\renewcommand{\arraystretch}{1.0} 
\begin{table*}[t]
	\centering
	\begin{tabular}{l|c||ccccc|c||ccc|c}
		\hline\hline
		\multirow{2}{*}{Method} &
		\multirow{2}{*}{Feat.} & 
		\multicolumn{6}{c||}{THUMOS14} & \multicolumn{4}{c}{ActivityNet-v1.3} \\
		\cline{3-12}
		& & $0.3$ & $0.4$ & $0.5$ & $0.6$ & $0.7$ & Avg. & $0.5$ & $0.75$ & $0.95$ & Avg. \\ \hline\hline
		BSN~\cite{lin2018bsn} & TSN & $53.5$ & $45.0$ & $36.9$ & $28.4$ & $20.0$ & $36.8$ & $46.46$ & $29.96$ & $8.02$ & $29.17$\\
		BMN~\cite{lin2019bmn} & TSN & $56.0$ & $47.4$ & $38.8$ & $29.7$ & $20.5$ & $38.5$ & $50.07$ & $34.78$ & $8.29$ & $33.85$ \\
		GTAD~\cite{xu2020gtad} & TSN & $54.5$ & $47.6$ & $40.2$ & $30.8$ & $23.4$ & $39.3$ & $50.36$ & $34.60$ & $9.02$ & $34.09$ \\
		MUSES~\cite{liu2021muse} & I3D & $68.9$ & $64.0$ & $56.9$ & $46.3$ & $31.0$ & $53.4$ & $50.02$ & $34.97$ & $6.57$ & $33.99$ \\
		PBRNet~\cite{liu2020progressive} & I3D & $58.5$ & $54.6$ & $51.3$ & $41.8$ & $29.5$ & $47.1$ & $53.96$ & $34.97$ & $8.98$ & $35.01$ \\
		RTD-Net~\cite{tan2021relaxed} & Mix & $68.3$ & $62.3$ & $51.9$ & $38.8$ & $23.7$ & $49.0$ & $47.21$ & $30.68$ & $8.61$ & $30.83$ \\
		VSGN~\cite{zhao2021stitching} & $*$ & $66.7$ & $60.4$ & $52.4$ & $41.0$ & $30.4$ & $50.2$ & $52.38$ & $36.01$ & $8.37$ & $35.07$ \\
		ContextLoc~\cite{zhu2021contextloc} & I3D & $68.3$ & $63.8$ & $54.3$ & $41.8$ & $26.2$ & $50.9$ & $56.01$ & $35.19$ & $3.55$ & $34.23$ \\
		AFSD~\cite{lin2021salient} & I3D & $67.3$ & $62.4$ & $55.5$ & $43.7$ & $31.1$ & $52.0$ & $52.40$ & $35.30$ & $6.50$ & $34.40$ \\
        DCAN & TSN & $68.2$ & $62.7$ & $54.1$ & $43.9$ & $32.6$ & $52.3$ & $51.78$ & $35.98$ & $9.45$ & $35.39$ \\
		Zhu \textit{et al.}~\cite{zhu2022disentangled} & I3D & $72.1$ & $65.9$ & $57.0$ & $44.2$ & $28.5$ & $53.5$ & $58.14$ & $36.30$ & $6.16$ & $35.24$ \\
		RCL~\cite{wang2022rcl} & * & $70.1$ & $62.3$ & $52.9$ & $42.7$ & $30.7$ & $51.0$ & $54.19$ & $36.19$ & $9.17$ & $35.98$ \\
		TAGS~\cite{nag2022tags} & I3D & $68.6$ & $63.8$ & $57.0$ & $46.3$ & $31.8$ & $52.8$ & $\boldsymbol{56.30}$ & $36.80$ & $9.60$ & $36.50$ \\
		TALLFormer~\cite{cheng2022tallformer} & Swin & $76.0$ & - & $63.2$ & - & $34.5$ & $59.2$ & $54.10$ & $36.20$ & $7.90$ & $35.60$ \\
		\hline
        FCOS~\cite{tian2019fcos} & I3D & $62.6$ & $56.3$ & $47.9$ & $36.5$ & $23.2$ & $45.3$ & $50.22$ & $33.51$ & $5.57$ & $32.30$ \\
         + BRN (Ours) & I3D & $71.7$ & $65.8$ & $55.5$ & $44.6$ & $29.4$ & $53.4$ & $52.41$ & $\boldsymbol{37.72}$ & $\boldsymbol{9.89}$ & $36.16$ \\
        \hline
        ActionFormer~\cite{zhang2022actionformer} & I3D & $\boldsymbol{82.1}$ & $77.8$ & $71.0$ & $59.4$ & $43.9$ & $66.8$ & $53.50$ & $36.20$ & $8.20$ & $35.60$ \\
         + BRN (Ours) & I3D & $82.0$ & $\boldsymbol{78.8}$ & $\boldsymbol{71.9}$ & $\boldsymbol{59.9}$ & $\boldsymbol{45.4}$ & $\boldsymbol{67.6}$ & $54.89$ & $37.50$ & $8.36$ & $\boldsymbol{36.69}$ \\
		\hline\hline
	\end{tabular}
	\caption{\textbf{The comparison results with the-state-of-the-art on THUMOS14 and ActivityNet-v1.3.} 
	`-' means that the score is not available. 
	`Avg.' is the average score at IoU thresholds. `Features' indicates the backbone network for feature extraction. `Mix' stands for use of TSN and I3D backbones together, and `$*$' means that we report the best performance among the backbones.}
	\label{tab:main}
\end{table*}
\endgroup

Table.~\ref{tab:main} shows the results compared with the-state-of-the-art methods in THUMOS14.
As we can see in the table, our framework outperforms the baselines in terms of the average score.
Especially for the FCOS backbone, the performance gain is significant.
We conjecture that our scale-time representations effectively restore many missed boundary features thus reaching a high recall performance.

The table also shows the results of comparison with the-state-of-the-art methods in ActivityNet-v1.3.
As can be seen in the table, our framework outperforms the baselines in $0.75$, $0.95$, and average IoUs.
Especially, for the score on the $0.95$ IoU threshold with the FCOS backbone, our framework shows the best score over all the baselines.
One of the possible reasons is that our scale convolutions in scale-time blocks aggregate features over all the scales including those from the finer-scale level. 
We expect that these fine-grained features improve the precision especially for long instances so that the APs at high IoUs are significantly boosted.

\subsection{Vanishing Boundary Problem}
\noindent\textbf{Performances of Neighboring Instances.}
We conduct experiments with neighboring ground-truth (GT) action instances on Activitynet-v1.3 based on FCOS to evaluate the degree of the problem.
We expect that the performance for instances which are temporally close to each other will increase when the problem is alleviated.
In the experiments, we first calculate the distance between a GT instance and its nearest neighbor GT instance, and further normalize it with the whole video length, called the distance ratio.
We use instances with distance ratio $\leq 0.50$ as neighboring instances.

Table.~\ref{tab:performances_of_neighboring_instances} shows the results.
Note that the ratios of $0.25$, and $0.50$ mean the instances subject to $\text{ratio} \le 0.25$, and $0.25 < \text{ratio} \le 0.50$, respectively.
We can see that our model outperforms the baselines on all the scores.
As for the ratio $0.25$, our framework shows a remarkable improvement compared to FCOS.
Although AFSD also shows an improved performance by deploying high-resolution features for refinement, our framework achieves a better performance when compared with the baseline models including AFSD.

\begingroup
\setlength{\tabcolsep}{12.5pt} 
\renewcommand{\arraystretch}{1.00} 
\begin{table}[t]
	\centering
	\begin{tabular}{l|cc|cc}
		\hline\hline
		\multirow{2}{*}{Method} & \multicolumn{2}{c|}{mAP ($\uparrow$)} & \multicolumn{2}{c}{FNR ($\downarrow$)} \\
		& XS & S & XS & S \\ 
		\hline\hline
		FCOS & $8.1$ & $24.5$ & $67.0$ & $48.7$ \\
		\hline
		BMN$^{\dagger}$~\cite{lin2019bmn} & $8.7$ & $24.4$ & $67.9$ & $44.4$ \\
		GTAD$^{\dagger}$~\cite{xu2020gtad} & $7.8$ & $24.2$ & $68.9$ & $44.0$\\
		AFSD~\cite{lin2021salient} & $9.7$ & $22.3$ & $65.9$ & $51.9$\\
		\hline
        BRN~(Ours) & $\boldsymbol{11.8}$ & $\boldsymbol{28.4}$ & $\boldsymbol{61.2}$ & $\boldsymbol{40.2}$ \\
		\hline\hline
	\end{tabular}
	\caption{\textbf{Performances of scales.} We conduct DETAD experiments on ActivityNet-v1.3 in small scales to validate our framework for the vanishing boundary problem. `$\dagger$' indicates reproduced models.}
	\label{tab:performances_of_scales}
\end{table}
\endgroup

\vspace{3pt}
\noindent\textbf{Performances of Scales.}
We point out that directly deploying FPN without consideration of the vanishing boundary problem can cause performance degradation.
In order to analyze the effect of our framework for solving the problem, 
we conduct experiments following DETAD~\cite{alwassel2018detad} with small scale groups based on their temporal scales: Extra Small (XS), and Small (S).
The groups are defined by the ratio of the GT instance length to the total video length, which is called `Coverage' in DETAD.
That is, extra small (XS) instances have the ratio less than or equal to $0.20$, small (S) ones have the ratio in $(0.20, 0.40]$ defined by DETAD.
The experiments are on ActivityNet-v1.3, and therefore, the ratio (Coverage) is more suitable than the absolute time length as the features are resized to a fixed length in inference.

Table.~\ref{tab:performances_of_scales} shows the mAP scores and false negative rates (FNR) of the two groups on ActivityNet-v1.3.
We compare ours with BMN, GTAD as single-scale models, and FCOS, AFSD as multi-scale baselines.
As we can see, the mAP scores for the extra small and small-scale instances are higher than those of the baselines.
More importantly, the false negative rates of ours are much lower than those of the baselines.
This means that our scale-time representations effectively relieve the problem of vanishing boundary so that more short instances are predicted correctly while suppressing longer false positives.

\begin{figure}[t]
\centering
\includegraphics[width=8.0cm]{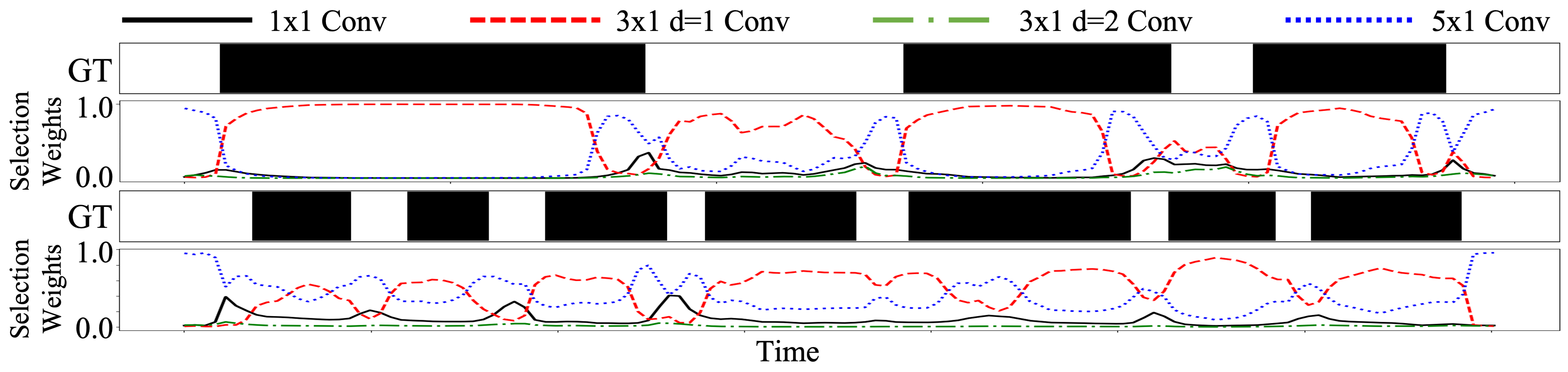}
\caption{\textbf{Examples of selection weights.} 
The figure contains illustration of two examples with the selection weights of the final scale convolution block on test samples in ActivityNet-v1.3.}
\label{fig:selection_weights}
\end{figure}

\vspace{3pt}
\noindent
\textbf{Visualization of Selection Weights.}
Our STBs are specifically designed to address the vanishing boundary problem by learning feature fusion.
STBs have two key characteristics: 1) feature enhancement and 2) feature correction.
Fig.~\ref{fig:selection_weights} shows that the selection weights of a kernel size of 3 are predominantly high within actions, enhancing features with neighboring feature levels.
Conversely, the weights of a size of 5 are elevated in boundary regions, allowing for the correction of missing boundary patterns from distant feature scales.
This illustrates how BRN handles the problem.

\subsection{Ablation Study}
\noindent\textbf{Scale Convolutions.}
In this study, we examine the effect of scale and time convolutions via ablating each component of STB.
Note that we use smaller models with the feature length of 128 for the ablation study with ActivityNet-v1.3.
In our framework, time convolutions are to mix and spread out the features from scale convolutions while scale convolutions aggregate the features over scales to cope with the vanishing boundary problem.

As shown in Table.~\ref{tab:scale_convolutions}, both of the scale and time convolutions improve the performance.
From the results, we can find that scale convolutions are the key part of BRN as the absence of scale convolutions significantly drops the overall performance.

\begingroup
\setlength{\tabcolsep}{12.5pt} 
\renewcommand{\arraystretch}{1.00} 
\begin{table}[t]
	\centering
	\begin{tabular}{l|cc|c}
		\hline\hline
		\multirow{2}{*}{Method} & $0.25$ & $0.50$ & \multirow{2}{*}{Avg.} \\
		& $(38.89\%)$ & $(10.95\%)$ & \\
		\hline\hline
		FCOS & $12.98$ & $21.69$ & $17.34$ \\
		\hline
		BMN$^{\dagger}$~\cite{lin2019bmn} & $12.85$ & $18.61$ & $15.73$ \\
		GTAD$^{\dagger}$~\cite{xu2020gtad} & $14.94$ & $19.42$ & $17.18$ \\
		AFSD~\cite{lin2021salient} & $15.90$ & $19.61$ & $17.76$ \\
		\hline
        BRN~(Ours) & $\boldsymbol{16.48}$ & $\boldsymbol{25.22}$ & $\boldsymbol{20.85}$ \\
		\hline\hline
	\end{tabular}
	\caption{\textbf{Performances of neighboring instances.} It shows performances of neighboring instances with distance ratios less than or equal to $0.50$ on ActivityNet-v1.3. `$\dagger$' indicates reproduced models.}
	\label{tab:performances_of_neighboring_instances}
\end{table}
\endgroup

\begin{table*}[t]
\begingroup
\setlength{\tabcolsep}{10.00pt} 
\renewcommand{\arraystretch}{1.00} 
\begin{subtable}[t]{0.49\textwidth}
	\centering
	\begin{tabular}{c|c|ccc|c}
		\hline\hline
		Scale & Time & $0.5$ & $0.75$ & $0.95$ & Avg. \\ \hline\hline
		$\cdot$ & $\cdot$ & $50.22$ & $33.51$ & $5.57$ & $32.30$ \\
        $\cdot$ & \checkmark & $50.65$ & $33.85$ & $5.80$ & $32.56$ \\
        \checkmark & $\cdot$ & $51.55$ & $36.69$ & $9.63$ & $35.58$ \\
        \checkmark & \checkmark & $51.58$ & $36.92$ & $9.42$ & $35.64$ \\
		\hline\hline
	\end{tabular}
	\caption{\textbf{Scale convolutions}. 
	To validate the effect of scale convolutions in STB, ablation study is conducted.}
	\label{tab:scale_convolutions}
\end{subtable}
\endgroup
\hfill
\begingroup
\setlength{\tabcolsep}{9.9pt} 
\renewcommand{\arraystretch}{1.00} 
\begin{subtable}[t]{0.49\textwidth}
	\centering
	\begin{tabular}{c|c|ccc|c}
		\hline\hline
		Select & Dilate & $0.5$ & $0.75$ & $0.95$ & Avg. \\ \hline\hline
        $\cdot$ & \checkmark & $50.83$ & $36.14$ & $9.35$ & $35.07$ \\
        \checkmark & $\cdot$ & $51.16$ & $36.60$ & $9.37$ & $35.32$ \\
        \checkmark & \checkmark & $51.58$ & $36.92$ & $9.42$ & $35.64$ \\
		\hline\hline
	\end{tabular}
	\caption{\textbf{Selection \& Dilation in STB}. 
	In order to validate the effect of the selection module and dilated convolutions in STB, we ablate each component.}
	\label{tab:selection_and_dilation}
\end{subtable}
\endgroup
\vfill
\begingroup
\setlength{\tabcolsep}{11.50pt} 
\renewcommand{\arraystretch}{1.00} 
\begin{subtable}[t]{0.49\textwidth}
	\centering
	\begin{tabular}{l|ccc|c}
		\hline\hline
		Method & $0.5$ & $0.75$ & $0.95$ & Avg. \\ \hline\hline
        Self-Attn. & $50.88$ & $36.55$ & $9.47$ & $35.20$ \\
        Convolution & $51.58$ & $36.92$ & $9.42$ & $35.64$ \\
		\hline\hline
	\end{tabular}
	\caption{\textbf{Scale aggregation}. 
	In order to fuse the features over scales, we can have different architectural choices such as self-attention.}
	\label{tab:scale_aggregation}
\end{subtable}
\endgroup
\hfill
\begingroup
\setlength{\tabcolsep}{12.40pt} 
\renewcommand{\arraystretch}{1.00} 
\begin{subtable}[t]{0.49\textwidth}
	\centering
	\begin{tabular}{l|ccc|c}
		\hline\hline
		Method & $0.5$ & $0.75$ & $0.95$ & Avg. \\ \hline\hline
        w/o Sep. & $51.50$ & $36.74$ & $10.00$ & $35.58$ \\
        w/~~~Sep. & $51.58$ & $36.92$ & $9.42$ & $35.64$ \\
		\hline\hline
	\end{tabular}
	\caption{\textbf{Separation of scale-time convolutions.}
	The table contains the results of ablation on the separation of scale-time convolutions.}
	\label{tab:separation_scale_time_convolutions}
\end{subtable}
\endgroup
\caption{\textbf{Ablation study.} Experiments of the ablation study are based on ActivityNet-v1.3 with the FCOS models of $T=128$.}
\end{table*}

\vspace{3pt}
\noindent
\textbf{Scale Aggregation.}
When it comes to feature exchange, we have more options on the architecture like self-attention~\cite{vaswani2017attention}.
Therefore, we also conduct experiments with self-attention modules as feature aggregation along the scale axis.
Table.~\ref{tab:scale_aggregation} shows the results of the experiments with the self-attention and scale-convolution methods.
We can find that scale convolutions show a higher performance than self-attention.
One of the possible reasons is that the multi-scale features have high similarity for neighboring levels, but self-attention modules do not fully exploit the property when fusing the features.

\vspace{3pt}
\noindent
\textbf{Selection \& Dilation in STB.}
In our scale-time blocks, we propose the selection module and dilation convolutions to handle the degree of exchanging features from distant scale levels.
To further analyze the benefits of each design, we averaged the outputs $O_i$ of the four kernels without learnable selection weights for ablating selection.
Also, we unify the rates of dilation as $1$ in STB for dilation.

As shown in Table.~\ref{tab:selection_and_dilation}, the ablation on them brings $0.57$ ($1.6$\%) and $0.32$ ($0.9\%$) drop in the average mAP.
In addition, convolutions with various receptive fields have strengths to handle degree of exchange features from distant scales, and improvements are not just from the increased number of layers.

\vspace{3pt}
\noindent
\textbf{Separation of Scale-Time Convolutions.}
In STBs, scale and time convolutions are stacked sequentially.
On the other hand, we can also construct these scale-time blocks in unified operations with 2D convolution kernels of $S \times T$.
Therefore, we have ablated separation of scale and time blocks in this study by merging the two convolutions in one convolution block with 2D $S \times T$ kernels.

The results are shown in Table.~\ref{tab:separation_scale_time_convolutions}.
The ablation on the separation brings a $0.06$ ($0.2$\%) drop in the mAP in spite of $50$\% more parameters for the scale-time convolution blocks.
The main reason for separation in our framework is that representations of scale and time are basically different, which are also aligned with the trend of separation between space and time in 3D CNN for video representations like R(2+1)D~\cite{tran2018closer}, S3D~\cite{xie2018s3d}, and SlowFast~\cite{feichtenhofer2018slowfast}.
Therefore, unifying the two different convolutions in one block could cause a performance degradation with additional computational overhead.

\begingroup
\setlength{\tabcolsep}{10.8pt} 
\renewcommand{\arraystretch}{1.00} 
\begin{table}[t]
	\centering
	\begin{tabular}{l|ccc|c}
		\hline\hline
		Method & $0.5$ & $0.75$ & $0.95$ & Avg. \\ \hline\hline
		No Dilation & $51.16$ & $36.60$ & $9.37$ & $35.32$ \\
		$3 \times 1$ Dilation & $51.69$ & $36.68$ & $9.48$ & $35.66$ \\
		\hline
		Ours & $51.58$ & $36.92$ & $9.42$ & $35.64$ \\
		\hline\hline
	\end{tabular}
	\caption{\textbf{Ablation study on dilation rates and kernel sizes in STB}. On top of the study on dilation, we have also conducted experiments on the convolution kernel sizes with the FCOS models of $T=128$.}
	\label{tab:dilation}
\end{table}
\endgroup

\vspace{2pt}\noindent
\textbf{Dilation Rates \& Kernel Sizes: }
In the paper, we have conducted an ablation study on dilated convolutions in our STB.
To further validate our design, we have conducted an ablation study on the kernel sizes.
As shown in Table.~\ref{tab:dilation}, unified dilation rates bring a performance drop of $0.32$ ($0.9\%$) in the averaged mAP.

Instead of using different sizes like 1, 3, or 5, we only use $3 \times 1$ but with dilation rates of $1$, $2$, $3$, and $4$, respectively, for each scale convolution layer.
Although the size of scale levels is $5$, it also shows a slight improvement than the proposed one of the paper as shown in '$3 \times 1$ Dilation' of the table.
This is possibly because it enlarges effective receptive fields like ASPP~\cite{chen2017deeplab} in semantic segmentation.
From this result, we can conclude that the most crucial part of improvement is the various receptive fields so that they can select the appropriate scale levels of features to exchange.

\begingroup
\setlength{\tabcolsep}{11.8pt} 
\renewcommand{\arraystretch}{1.0} 
\begin{table}[t]
	\centering
	\begin{tabular}{l|ccc|c}
		\hline\hline
		\# of STB & $0.5$ & $0.75$ & $0.95$ & Avg. \\ \hline\hline
        N=1 & $51.65$ & $36.83$ & $9.18$ & $35.51$ \\
        N=2 & $51.66$ & $36.95$ & $9.66$ & $35.66$ \\
        N=3 & $51.58$ & $36.92$ & $9.42$ & $35.64$ \\
        N=4 & $51.89$ & $36.84$ & $9.37$ & $35.68$ \\
        N=5 & $51.78$ & $36.65$ & $9.73$ & $35.58$ \\
		\hline\hline
	\end{tabular}
	\caption{\textbf{Scale-time blocks}. 
	The table shows the results of the experiments by varying the number of scale-time blocks (STB).
	The study is conducted on ActivityNet-v1.3 with the FCOS models of $T=128$.
	}
	\label{tab:scale_time_block}
\end{table}
\endgroup

\vspace{2pt}\noindent
\textbf{Stacking Scale-Time Blocks.} 
The scale-time blocks (STB) of our framework can be stacked multiple times.
Hence, we have also investigated the effects of the number of scale-time blocks.

As reported in Table.~\ref{tab:scale_time_block}, 
all configurations significantly improve the overall performance.
Interestingly, only one block of the scale-time block remarkably improves the performance with less parameters.
We use the $N=3$ configuration as the default model.

\vspace{2.0pt}\noindent
\textbf{Differences b/w Scale-Time Representations and FPN.} 
One might point out that our scale-time representations could make a similar effect to feature pyramid network (FPN)~\cite{lin2017fpn} because we also use interpolation followed by convolutions.
However, there are two main differences between ours and FPN.

First, our scale-time representations construct a new axis called `scale'.
This enables the model to learn direct feature fusion on the scale axis with convolutions.
We can observe that features from neighboring scale levels exhibit similar patterns, which are aligned with the basic heuristics for convolutions in images.
Therefore, convolutions on the scale axis are a lot more effective for aggregating the features especially over distant levels than FPN.
More importantly, FPN operates only on the features of neighboring scale levels while our scale-time convolutions can focus more on the features from distant scale levels.

\begin{figure}[t]
\centering
\includegraphics[width=8.3cm]{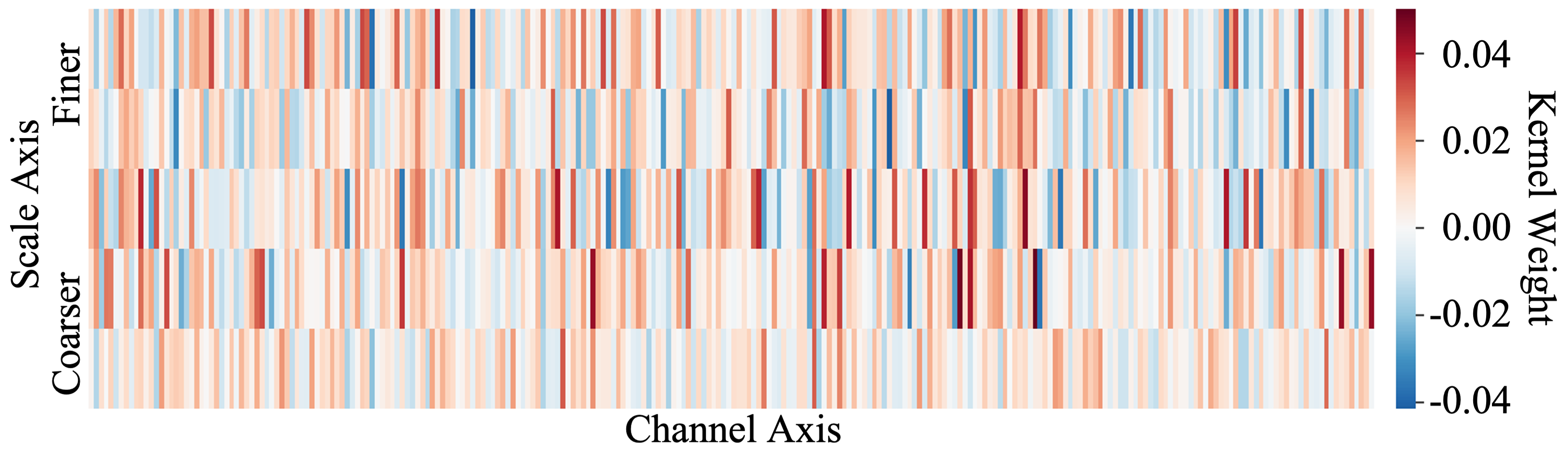}
\caption{\textbf{Illustration of a scale convolution kernel.} The figure shows the learned $5 \times 1$ scale convolution kernel in the final scale convolution block. Note that visualization is for one output channel.}
\label{fig:scale_conv_kernel}
\end{figure}

Second, 
Fig.~\ref{fig:scale_conv_kernel} shows the learned $5 \times 1$ scale convolution kernel for an output channel in the final scale convolution block.
As shown, the kernel weights are varied for each input channel since some channels do not need to be shared from distant scale levels.
This variety is also found in the kernels of size $3$.
This can be one of the main differences between ours and FPN.
Scale convolutions are a more direct way to fuse features over scales, and more importantly, it can exchange features from distant levels differently weighing the channels while FPN fuses features in the same way over channels.

\begin{figure*}[t]
\centering
\includegraphics[width=16.00cm]{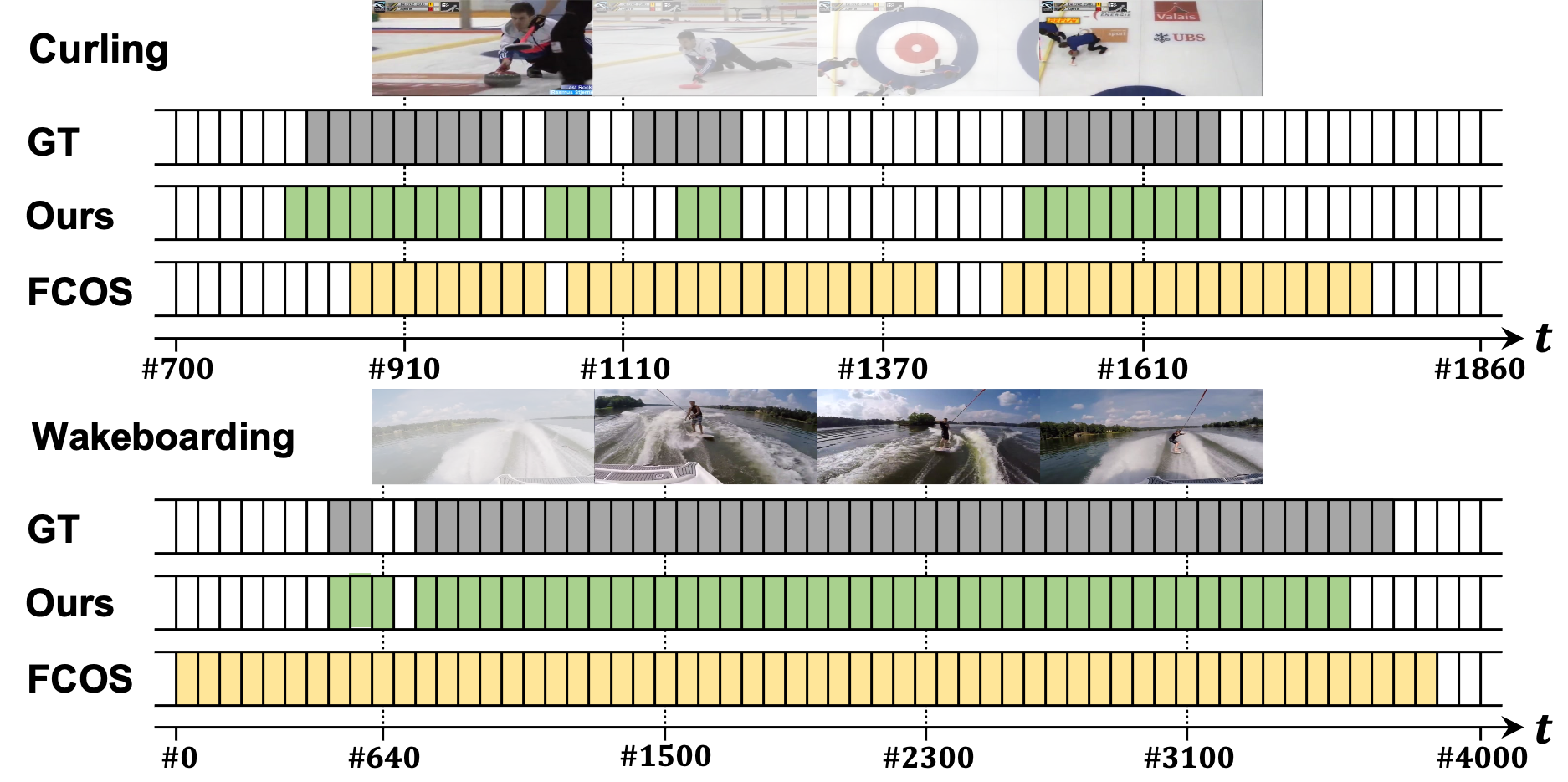}
\caption{\textbf{Visualization related to the vanishing boundary problem.} The figure shows samples from the validation set of ActivityNet-v1.3. As seen, FCOS produces longer false positives for neighboring short instances due to the vanishing boundary problem. However, our model of the FCOS backbone precisely localizes them.
}
\label{fig:visualization}
\end{figure*}

\vspace{2pt}\noindent
\textbf{Visualization of Neighboring Instances.}
Fig.~\ref{fig:visualization} shows visualization of samples from the validation set of ActivityNet-v1.3.
As shown, short neighboring instances are falsely considered as a long one in FCOS due to the vanishing boundary problem.
However, our model localizes each instance precisely.

\vspace{2pt}\noindent
\textbf{Different Trends b/w Two Datasets.}
Our model shows more improvements in lower IoUs for THUMOS14, a different trend of ActivityNet.
We conjecture that it is mainly from different length distributions of action instances.

\begin{figure}[t]
\centering
\includegraphics[width=8.3cm]{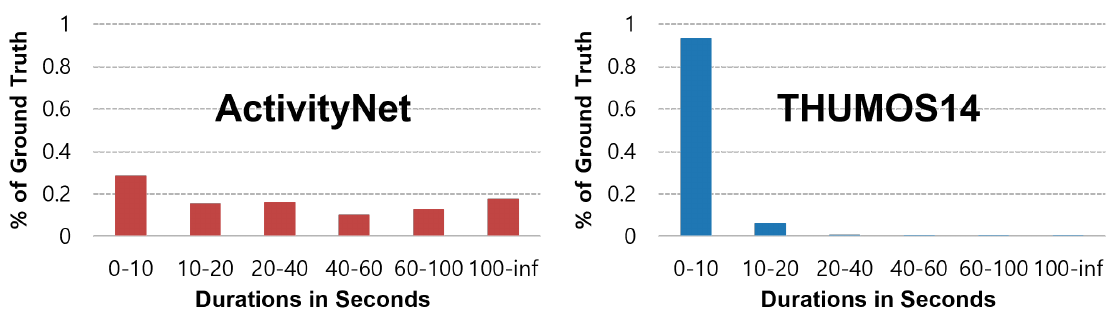}
\caption{\textbf{Temporal length distributions of instances in two benchmarks.}
To analyze the different trend of performances on THUMOS14 and ActivityNet-v1.3, we have calculated statistics of time lengths for the two benchmarks.
}
\label{fig:length_distributions}
\end{figure}

THUMOS14 has a lot more short action instances than ActivityNet as shown in Fig.~\ref{fig:length_distributions} where we report absolute duration as video lengths of two datasets are comparable.
Specifically, THUMOS14 has the 110-frame action length in average, but ActivityNet has the 936 frames.

As known, it is harder to achieve a high IoU for short instances than long ones.
BRN shows a higher performance for short instances than baselines by relieving VBP.
That is, BRN shows 34.13 in the average mAP for short instances ($<5s$) of THUMOS14, but AFSD~\cite{lin2021salient} has 29.71 although performances for other scales are comparable.
In this sense, we expect that the high performance in short instances more boosts performances at relatively lower IoUs.


%% file: sec/5_conclusion.tex
\section{Conclusion}
In this paper, we have proposed Boundary-Recovering Network (BRN) with scale-time representations to relieve the vanishing boundary problem.
Scale-time blocks allow the model to learn aggregation of features over scales while preserving crucial boundary representations.
Through our extensive experiments, BRN has shown the-state-of-the-art performances on the challenging benchmarks.
More importantly, we have demonstrated that our model significantly improves the performance of neighboring instances with more balanced accuracies over diverse scales through explicitly alleviating the problem.
In this way, we believe that BRN can serve as a strong multi-scale architecture for TAD.